# A Comparative Study of Adaptive Crossover Operators for Genetic Algorithms to Resolve the Traveling Salesman Problem

ABDOUN Otman
LaRIT, Department of Computer Science
IBN Tofail University, Kenitra, Morocco

ABOUCHABAKA Jaafar
LaRIT, Department of Computer Science
IBN Tofail University, Kenitra, Morocco

## ABSTRACT
Genetic algorithm includes some parameters that should be adjusting so that the algorithm can provide positive results. Crossover operators play very important role by constructing competitive Genetic Algorithms (GAs). In this paper, the basic conceptual features and specific characteristics of various crossover operators in the context of the Traveling Salesman Problem (TSP) are discussed. The results of experimental comparison of more than six different crossover operators for the TSP are presented. The experiment results show that OX operator enables to achieve a better solutions than other operators tested.

## Keywords
Travelers Salesman Problem, Genetic Algorithm, NP-Hard Problem, Crossover Operator, probability of crossover, Genetic Algorithm,

## 1. INTRODUCTION
This section introduces the current scientific understanding of the natural selection process with the purpose of gaining an insight into the construction, application, and terminology of genetic algorithms. Natural selection –evolution- is discussed in many texts and treatises, and one of its first proponents, Charles Darwin.His theory of evolution was based on four primary premises [7]. First, like begets like; equivalently, an offspring has many of the characteristics of its parents. This premise implies that the population is stable. Second, there are variations in characteristics between individuals that can be passed from one generation to the next. The third premise is that only a small percentage of the offspring produced survive to adulthood. Finally, which of the offspring survive depends on their inherited characteristics. These premises combine to produce the theory of natural selection. In modern evolutionary theory an understanding of genetics adds impetus to the explanation of the stages of natural selection.

Another set of biologically-inspired methods are Genetic Algorithms (GAs). They derive their inspiration from combining the concept of genetic recombination with the theory of evolution and survival of the fittest members of a population [5]. Starting from a random set of candidate parameters, the learning process devises better and better approximations to the optimal parameters. The GA is primarily a search and optimization technique. One can, however, pose nearly any practical problem as one of optimization, including many environmental modeling problems. To configure a problem for GA solution requires that the modeler not only choose the representation methodology, but also the cost function that judges the model's soundness.
The genetic algorithm is a one of the family of evolutionary algorithms. The population of a genetic algorithm (GA) evolves by using genetic operators inspired by the evolutionary in biology, **"***The survival is the individual most suitable to the environment***"**. Darwin discovered that species evolution based on two components: the selection and reproduction. The selection provides a reproduction of the strongest and more robust individuals, while the reproduction is a phase in which the evolution run.

Genetic algorithms are powerful methods of optimization used successfully in different problems. Their performance is depending on the encoding scheme and the choice of genetic operators especially, the selection, crossover and mutation operators. A variety of these latest operators have been suggested in the previous researches. In particular, several crossover operators have been developed and adapted to the permutation presentations that can be used in a large variety of combinatorial optimization problems. In this area, a typical example of the most studied problems is the Traveling Salesman Problem (TSP).

The traveling salesman problem (TSP) is a classical problem of combinatorial optimization of Operations Research's area. The purpose is to find a minimum total cost Hamiltonian cycle [22]. There are several practical uses for this problem, such as vehicle routing (with the additional constraints of vehicle's route, such as capacity's vehicles) [23] and drilling problems [24].

The TSP has received considerable attention over the last two decades and various approaches are proposed to solve the problem, such as branch-and-bound [28], cutting planes [35], 2-opt [33], simulated annealing [31], neural network [1,37], and tabu search [9, 29]. Some of these methods are exact algorithms, while the others are near-optimal or approximate algorithms. The exact algorithms include the integer linear programming approaches with additional linear constraints to eliminate infeasible subtours [25, 27, 30, 34, 36,36]. On the other hand, network models yield appropriate methods that are flexible enough to include the precedence constraints [28,32]. More recently, genetic algorithm (GA) approaches are successfully implemented to the TSP [26]. Potvin [35] presents survey of GA approaches for the general TSP.

These researches have provided the birth of several genetic mechanisms in particular, the selection, crossover and the mutation operators. In order to resolve the TSP problem, we propose in this paper to study empirically the impact affiliation of the different crossover operators.Finally we analyze the experimental results.

## 2. TRAVELING SALESMAN PROBLEM

The Traveling Salesman Problem (TSP) is one of the most intensively studied problems in computational mathematics.In the TSP problem, which is closely related to the Hamiltonian cycle problem, a salesman must visit n cities. Modeling the problem as a complete graph with n vertices, we can say that the salesman wishes to make a tour, or Hamiltonian cycle, visiting each city exactly once and finishing at the city he starts from [1]. Given the cost of travel between all cities, how should he plan his itinerary for minimum total cost of the entire tour?

As a concrete example, consider a delivery company with a central depot. Each day, it loads up each delivery truck at the depot and sends it around to deliver goods to several addresses. At the end of the day, each truck must end up back at the depotso that it is ready to be loaded for the next day. To reduce costs, the company wants to select an order of delivery stops that yields the lowest overall distance traveled by each truck. This problem is the well-known "Traveling Salesman Problem," andit is NP-complete [1]. It has no known efficient algorithm. Under certain assumptions, however, we know of efficient algorithms that give an overall distance which is not too far above the smallest possible.

The search space for the TSP is a set of permutations of n cities. Any single permutation of n cities yields a solution (which is a complete tour of n cities). The optimal solution is a permutation which yields the minimum cost of the tour. The size of the search space is n!.

In other words, a TSP of size V is defined by a set of points v= {v1, v2, …,vn} which vi a city marked by coordinates vi.x and vi.y where we define a metric distance function f as in (1). A solution of TSP problem is a form of scheduling T=(T[1],T[2],……,T[n], T[1]) which T[i] is a permutation on the set {1, 2, …,V}. The evaluation function calculates the adaptation of each solution of the problem by the following formula:

$$f = \sum_{i=1}^{n-1} \sqrt{(v_i.x - v_{i+1}.x)^2 + (v_i.y - v_{i+1}.y)^2}$$
$$+ \sqrt{(v_n.x - v_1.x)^2 + (v_n.y - v_1.y)^2} \quad (1)$$

Where $n$ is the number of cities.

If $d$, a distance matrix, is added to the TSP problem, and $d(i,j)$ a distance between the city $v_i$ and $v_j$ (2), hence the cost function $f$ (1) can be expressed as follows:

$$d(i,j) = \sqrt{(v_i.x - v_j.x)^2 + (v_i.y - v_j.y)^2} \quad (2)$$

$$f(T) = \sum_{i=1}^{n-1} d(T[i], T[i+1]) + d(T[n], T[1]) \quad (3)$$

The mathematical formulation of TSP problem expresses by:

$$min\{f(T), T = (T[1], T[2], …… , T[n])\} \quad (4)$$

Which T[i] is a permutation on the set {1, 2, …,V}.

The travelling salesman problem (TSP) is an NP-hard problem in combinatorial optimization studied in operations research and theoretical computer science [5].

***Theorem:*** The subset-sum problem is NP-complete [3].

***Proof :***We first show that TSP belongs to NP. Given an instance of the problem, we use as a certificate the sequence of n vertices in the tour. The verification algorithm checks that this sequence contains each vertex exactly once, sums up the edge costs, and checks whether the sum is at most k. This process can certainly be done in polynomial time.

To prove that TSP is NP-hard, we show that ***HAM-CYCLE ≤ P*** TSP. Let $G =(V, E)$be an instance of HAM-CYCLE. We construct an instance of TSP asfollows. We form the complete graph $G' = (V, E')$, , where$E'=\{(i,j) : i, j \in V$ and$i \neq j \}$, and we define the cost function $c$ by

$$c(i,j) = \begin{cases} 0 & if (i,j) \in E \\ 1 & if (i,j) \notin E \end{cases} \quad (5)$$

(Note that because $G$is undirected, it has no self-loops, and so $c(v, v)=1$ for all vertices $v \in V$.) The instance of TSP is then $(G', c, 0)$, which we can easily create in polynomial time.
We now show that graph $G$has a Hamiltonian cycle if and only if graph$G'$has atour of cost at most$0$. Suppose that graph$G$has a Hamiltonian cycle$h$. Each edgein$h$belongs to$E$ and thus has cost$0$ in $G'$. Thus,$h$is a tour in$G'$with cost$0$.
Conversely, suppose that graph$G'$has a tour$h'$of cost at most$0$. Since the costsof the edges in$E'$are$0$ and$1$, the cost of tour$h'$is exactly$0$and each edge on thetour must have cost$0$. Therefore,$h'$contains only edges in$E$. We conclude that$h'$is a Hamiltonian cycle in graph$G$.

A quick calculation shows that the complexity is $O(n!)$ which n is the number of cities (Table. 1).

**Table 1. Number of possibilities and calculation time by the number of cities**

| Number of cities | Number of possibilities | Computation time |
|---|---|---|
| 5 | 12 | 12 μs |
| 10 | 181440 | 0,18 ms |
| 15 | 43 billions | 12 hours |
| 20 | 60 E+15 | 1928 years |
| 25 | 310 E+21 | 9,8 billions of years |

To solve the TSP, there are algorithms in the literature deterministic (exact) and approximation algorithms (heuristics).

## 2.1 Deterministic algorithm

During the last decades, several algorithms emerged to approximate the optimal solution: nearest neighbor, greedy algorithm, nearest insertion, farthest insertion, double minimum spanning tree, strip, space-filling curve and Karp, Litke and Christofides algorithm, etc. (some of these algorithms assume that the cities correspond to points in the plane under some standard metric).

The TSP can be modeled in a linear programming problem under constraints, as follows:

We associate to each city a number between 1 and V. For each pair of cities (i, j), we define $c_{ij}$ the transition cost from city $i$ to the city $j$, and the binary variable:

$$x_{ij} = \begin{cases} 1 & If the traveler moves from city i to city j \\ 0 & else \end{cases} \quad (6)$$

So the TSP problem can be formulated as a problem of integer linear programming, as follows:

$$min \sum_{i=1}^{n} \sum_{j=1}^{i-1} c_{ij} \ x_{ij} \quad (7)$$

Under the following constraints:

$$1 - \sum_{i\,j} x_{ij} = 2, \forall i \in N = \{1,2,\dots,n\} \quad (8)$$

$$2 - \sum_{i\in S} \sum_{j \notin S} x_{ij} \geq 2 \ for each S \subset N \quad (9)$$

There are several deterministic algorithms; we mention the method of separation and evaluation and the method of cutting planes.

The deterministic algorithm used to find the optimal solution, but its complexity is exponential order, and it takes a lot of memory space and it requires a very high computation time. In large size problems, this algorithm cannot be used.

Because of the complexity of the problem and the limitations of the linear programming approach, other approaches are needed.

## 2.2 Approximation algorithm

Many problems of practical significance are NP-complete, yet they are too important to abandon merely because we don't know how to find an optimal solution in polynomial time. Even if a problem is NP-complete, there may be hope. We have at least three ways to get around NP-completeness. First, if the actual inputs are small, an algorithm with exponential running time may be perfectly satisfactory. Second, we may be able to isolate important special cases that we can solve in polynomial time. Third, we might come up with approaches to find near-optimal solutions in polynomial time (either in the worst case or the expected case). In practice, near-optimality is often good enough. We call an algorithm that returns near-optimal solutions an *approximation algorithm*.

An approximate algorithm, like the Genetic Algorithms, Ant Colony [17] and Tabu Search [9], is a way of dealing with NP-completeness for optimization problem. This technique does not guarantee the best solution. The goal of an approximation algorithm is to come as close as possible to the optimum value in a reasonable amount of time which is at most polynomial time.

## 3. GENETIC ALGORITHM

A genetic algorithm (GA) is one such versatile optimization method. Figure 1 shows the optimization process of a GA – the two primary operations are mating and mutation. The GA combines the best of the last generation through mating, in which parameter values are exchanged between parents to form offspring. Some of the parameters mutate [6]. The objective function then judges the fitness of the new sets of parameters and the algorithm iterates until it converges. With these two operators, the GA is able to explore the full cost surface in order to avoid falling into local minima. At the same time, it exploits the best features of the last generation to converge to increasingly better parameter sets.

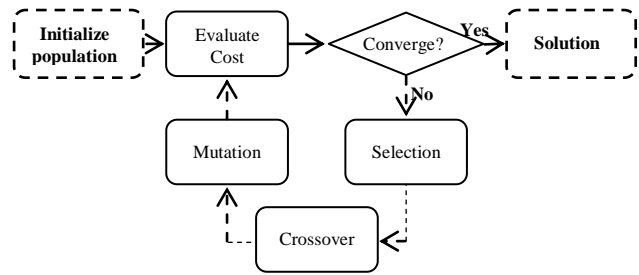

**Fig.1. Flowchart of optimization with a genetic algorithm**

GAs are remarkably robust and have been shown to solve difficult optimization problems that more traditional methods can not. Some of the advantages of GAs include:

- They are able to optimize disparate variables, whether they are inputs to analytic functions, experimental data, or numerical model output.

- They can optimize either real valued, binary variables, or integer variables.

- They can process a large number of variables.

- They can produce a list of best variables as well as the single best solution.

- They are good at finding a global minimum rather than local minima.

- They can simultaneously sample various portions of a cost surface.

- They are easily adapted to parallel computation.

Some disadvantages are the lack of viable convergence proofs and the fact that they are not known for their speed. As seen later in this chapter, speed can be gained by careful choice of GA parameters. Although mathematicians are concerned with convergence, often scientists and engineers are more interested in using a tool to find a better solution than obtained by other means. The GA is such a tool.

These algorithms were modeled on the natural evolution of species. We add to this evolution concepts the observed properties of genetics (Selection, Crossover, Mutation, etc), from which the name Genetic Algorithm. They attracted the interest of many researchers, starting with Holland [15], who developed the basic principles of genetic algorithm, and Goldberg [8] has used these principles to solve a specific optimization problems. Other researchers have followed this path [10]-[14].

## 3.1 Principles and Functioning

Irrespective of the problems treated, genetic algorithms, presented in figure (Fig. 1), are based on six principles:

- Each treated problem has a specific way to encode the individuals of the genetic population. A chromosome (a particular solution) has different ways of being coded: numeric, symbolic, or alphanumeric;

- Creation of an initial population formed by a finite number of solutions;

- Definition of an evaluation function (fitness) to evaluate a solution;

- Selection mechanism to generate new solutions, used to identify individuals in a population that could be crossed, there are several methods in the literature, citing the method of selection by rank, roulette, by tournament, random selection, etc.;

- Reproduce the new individuals by using Genetic operators:
    i. *Crossover operator:* is a genetic operator that combines two chromosomes (parents) to produce a new chromosome (children) with crossover probability $P_x$ ;
    ii. *Mutation operator:* it avoids establishing a uniform population unable to evolve. This operator used to modify the genes of a chromosome selected with a mutation probability $P_m$;

- Insertion mechanism: to decide who should stay and who should disappear.

- Stopping test: to make sure about the optimality of the solution obtained by the genetic algorithm.

We presented the various steps which constitute the general structure of a genetic algorithm: *Coding, method of selection, crossover and mutation operator and their probabilities, insertion mechanism, and the stopping test*. For each of these steps, there are several possibilities. The choice between these various possibilities allows us to create several variants of genetic algorithm. Subsequently, our work focuses on finding a solution to that combinative problem: What are the best settings which create an efficient genetic variant to solve the Traveling Salesman Problem?

## 4. APPLIED GENETIC ALGORITHMS TO THE TRAVELING SALESMAN PROBLEM

## 4.1 Problem representation methods

In this section we will present the most adapted method of data representation, the *path representation method*, with the treated problem.

The path representation is perhaps the most natural representation of a tour. A tour is encoded by an array of integers representing the successor and predecessor of each city.

**Table 2. Coding of a tour (3, 5, 2, 9, 7, 6, 8, 4)**

| 3 | 5 | 2 | 9 | 7 | 6 | 8 | 4 |
|---|---|---|---|---|---|---|---|

## 4.2 Generation of the initial population

The initial population conditions the speed and the convergence of the algorithm. For this, we applied several methods to generate the initial population:

- Random generation of the initial population.

- Generation of the first individual randomly, this one will be mutated **N-1** times with a mutation operator.

Generation of the first individual by using a heuristic mechanism. The successor of the first city is located at a distance smaller compared to the others. Next, we use a mutation operator on the route obtained in order to generate **(N-2)** other individuals who will constitute the initial population.

## 4.3 Selection

While there are many different types of selection, we will cover the most common type - roulette wheel selection. In roulette wheel selection, the individuals are given a probability $P_i$ of being selected (10) that is directly proportionate to their fitness. The algorithm for a roulette wheel selection algorithm is illustrated in algorithm (Fig. 3)

$$\frac{1}{N-1} \times \left( 1 - \frac{f_i}{\sum_{j \in Population} f_j} \right)(10)$$

Which $f_i$ is value of fitness function for the individual $i$.

```
for all members of population
    sum += fitness of this individual
endfor

for all members of population
    probability = sum of probabilities + (fitness / sum)
    sum of probabilities += probability
endfor

    number = Random between 0 and 1
for all members of population
if number > probability but less than next probability
then you have been selected
endfor
```

**Fig.2. Roulette wheel selection algorithm**

Thus, individuals who have low values of the fitness function may have a high chance of being selected among the individuals to cross.

## 4.4 Crossover Operator

The search of the solution space is done by creating new chromosomes from old ones. The most important search process is crossover. Firstly, a pair of parents is randomly selected from the mating pool. Secondly, a point, called crossover site, along their common length is randomly selected, and the information after the crossover site of the two parent strings are swapped, thus creating two new children. Of course, this basic crossover method does not support for the TSP [18]. The two newborn chromosomes may be better than their parents and the evolution process may continue. The crossover in carried out according to the crossover probability $P_x$.In this paper, we chose five crossover operators; we will explain their ways of proceeding in the following.

### 4.4.1  Uniform crossover operator

The child is formed by a alternating randomly between the two parents.

### 4.4.2  Cycle Crossover

The Cycle Crossover (CX) proposed by Oliver [15] builds offspring in such a way that each city (and its position) comes from one of the parents. We explain the mechanism of the cycle crossover using the following algorithm (Fig.3).

**Table 3. Cycle Crossover operator**

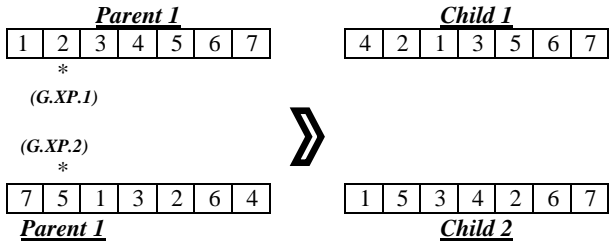

*Input:* Parents $x_1=[x_{1,1},x_{1,2},\ldots\ldots,x_{1,n}]$ and $x_2=[x_{2,1},x_{2,2},\ldots\ldots,x_{2,n}]$
*Output:* Children $y_1=[y_{1,1},y_{1,2},\ldots\ldots,y_{1,n}]$ and $y_2=[y_{2,1},y_{2,2},\ldots\ldots,y_{2,n}]$

--------------------------------------------------------------------------------
*Initialize*
- Initialize y1 and y2 being a empty genotypes;

$y_{1,1}=x_{1,1}$;
$y_{2,1}=x_{2,1}$;
i = 1;
**Repeat**
    j ← Index where we find $x_{2,i}$, in $X_1$;
    $y_{1,j}=x_{1,j}$;
    $y_{2,j}=x_{2,j}$;
    i = j;
**Until**  $x_{2,i}\notin y_1$

**For** each gene not yet initialized **do**
    **$y_{1,i}=x_{2,i}$;**
    **$y_{2,i}=x_{1,i}$;**
**Endfor**

**Fig.3. Cycle Crossover (CX) algorithm**

### 4.4.3  Partially-Mapped Crossover (PMX)

Partially matched crossover PMX noted, introduced by Goldberg and Lingel [19], is made by randomly choosing two crossover points XP1 and XP2 which break the two parents in three sections.

**Table 4. The partition of a parent**

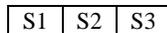

S1 and S3 the sequences of Parent1 are copied to the Child1, the sequence S2 of the Child1 is formed by the genes of Parent2, beginning with the start of its part S2 and leaping the genes that are already established. The algorithm (Fig.4) shows the crossover method PMX.

**Table 5. Example of PMX operator**

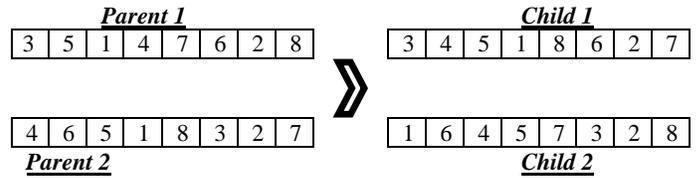

*Input:* Parents $x_1=[x_{1,1},x_{1,2},\ldots\ldots,x_{1,n}]$ and $x_2=[x_{2,1},x_{2,2},\ldots\ldots,x_{2,n}]$
*Output:* Children $y_1=[y_{1,1},y_{1,2},\ldots\ldots,y_{1,n}]$ and $y_2=[y_{2,1},y_{2,2},\ldots\ldots,y_{2,n}]$

--------------------------------------------------------------------------------
*Initialize*
- $y_1 = x_1$ and $y_2 = x_2$;
- Initialize $p_1$ and $p_2$ the position of each index in $y_1$ and $y_2$;
- Choose two crossover points a and b such that $1 \leq a \leq b \leq n$;

**for** each i between a and b do
    $t_1 = y_{1,i}$   and   $t_2 = y_{2,i}$;
    $y_{1,i} = t_2$   and   $y_{1,p1,t1} = t_1$;
    $y_{2,i} = t_1$   and   $y_{2,p2,t2} = t_2$;
    $p_{1,t1} = p_{1,t2}$  and  $p_{1,t2} = p_{1,t1}$;
    $p_{2,t1} = p_{2,t2}$  and  $p_{2,t2} = p_{2,t1}$;
**endfor**

**Fig.4. PMXCrossover Algorithm**

### 4.4.4  The uniform partially-mapped crossover (UPMX)

The Uniform Partially Matched Crossover presented by Cicirello and Smith [21], uses the technique of PMX. Any times, it does not use the crossover points; instead, it uses a probability of correspondence for each iteration. The algorithm (Fig.5) and the following example describe this crossover method.

**Table 6. UPMX operator example**

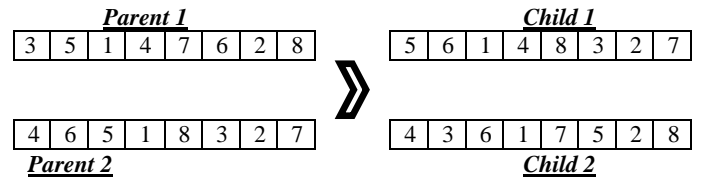

*Input:* Parents $x_1=[x_{1,1},x_{1,2},\ldots\ldots,x_{1,n}]$ and $x_2=[x_{2,1},x_{2,2},\ldots\ldots,x_{2,n}]$
*Output:* Children $y_1=[y_{1,1},y_{1,2},\ldots\ldots,y_{1,n}]$ and $y_2=[y_{2,1},y_{2,2},\ldots\ldots,y_{2,n}]$

--------------------------------------------------------------------------------
*Initialize*
- $y_1 = x_1$ and $y_2 = x_2$;
- Initialize $p_1$ and $p_2$ the position of each index in $y_1$ and $y_2$;
- Choose two crossover points a and b such that $1 \leq a \leq b \leq n$;

**For** each i between 1 and n **do**
    Chose a random number q between 0 and 1;
    **if** q ≥ p **then**
        $t_1 = y_{1,i}$   and   $t_2 = y_{2,i}$;
        $y_{1,i} = t_2$   and   $y_{1,p1,t1} = t_1$;
        $y_{2,i} = t_1$   and   $y_{2,p2,t2} = t_2$;
        $p_{1,t1} = p_{1,t2}$  and  $p_{1,t2} = p_{1,t1}$;
        $p_{2,t1} = p_{2,t2}$  and  $p_{2,t2} = p_{2,t1}$;
    **endif**
**endfor**

**Fig.5. Algorithm of UPMXCrossover**

### 4.4.5 Non-Wrapping Ordered Crossover (NWOX)

Non-Wrapping Ordered Crossover (NWOX) operator introduced by Cicirello [20], is based upon the principle of creating and filling holes, while keeping the absolute order of genes of individuals. The holes are created at the retranscription of the genotype, if $xj,i \in \{xk,a, \ldots ,xk,b\}$ then $xj,i$ is a hole. The example (Table.7) and the algorithm (Fig.6) explain this technique:

**Table 7. NWOX operator example**

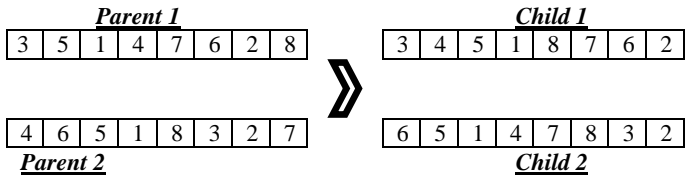

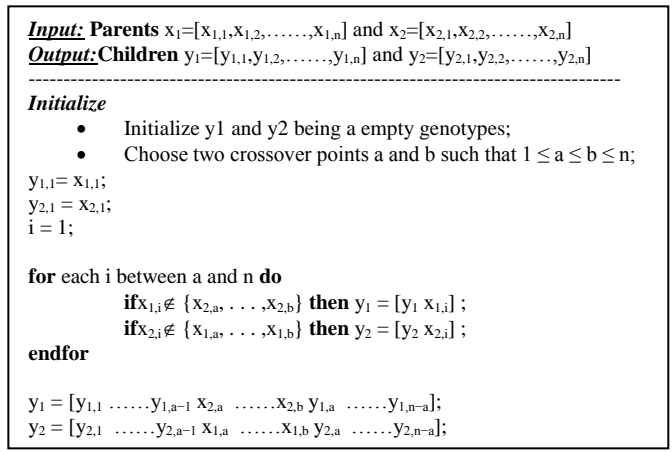

**Fig 6.Algorithm of NWOXcrossover operator**

### 4.4.6 Ordered Crossover (OX)

The Ordered Crossover method is presented by Goldberg[8], is used when the problem is of order based, for example in U-shaped assembly line balancing etc. Given two parent chromosomes, two random crossover points are selected partitioning them into a left, middle and right portion. The ordered two-point crossover behaves in the following way: child1 inherits its left and right section from parent1, and its middle section is determined.

**Table 8. OX operator example**

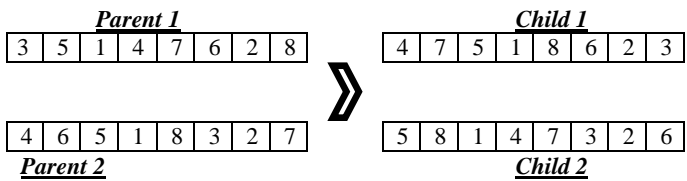

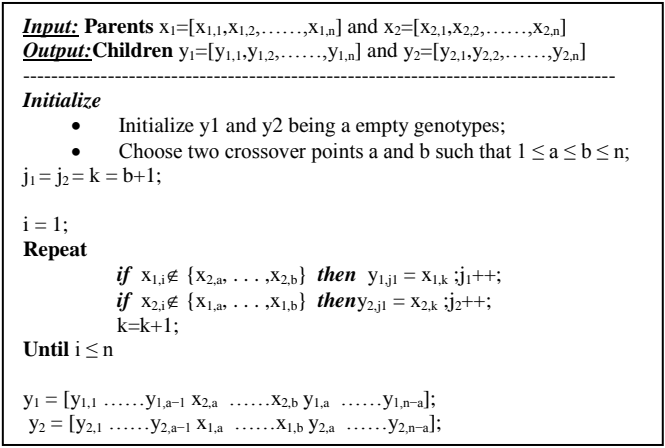

**Fig 7. Algorithm of Crossover operator OX**

### 4.4.7 Crossover with reduced surrogate

The reduced surrogate operator constrains crossover to always produce new individuals wherever possible. This is implemented by restricting the location of crossover points such that crossover points only occur where gene values differ.

### 4.4.8 Shuffle crossover

Shuffle crossover is related to uniform crossover. A single crossover position (as in single-point crossover) is selected. But before the variables are exchanged, they are randomly shuffled in both parents. After recombination, the variables in the offspring are unstuffed. This removes positional bias as the variables are randomly reassigned each time crossover is performed.

## 4.5 Mutation Operators

The two individuals (children) resulting from each crossover operation will now be subjected to the mutation operator in the final step to forming the new generation. This operator randomly flips or alters one or more bit values at randomly selected locations in a chromosome.

The mutation operator enhances the ability of the GA to find a near optimal solution to a given problem by maintaining a sufficient level of genetic variety in the population, which is needed to make sure that the entire solution space is used in the search for the best solution. In a sense, it serves as an insurance policy; it helps prevent the loss of genetic material.

In this study, we chose as mutation operator the Mutation methodReverse Sequence Mutation (RSM).

In the reverse sequence mutation operator, we take a sequence S limited by two positions i and j randomly chosen, such that i<j. The gene order in this sequence will be reversed by the same way as what has been covered in the previous operation. The algorithm (Fig. 8) shows the implementation of this mutation operator.

**Table 9.Mutation operator RSM**

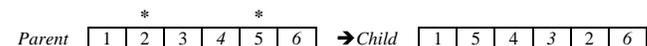

---

**Input:** Parents $x_1=[x_{1,1},x_{1,2},\ldots\ldots,x_{1,n}]$ and $x_2=[x_{2,1},x_{2,2},\ldots\ldots,x_{2,n}]$
**Output:** Children $y_1=[y_{1,1},y_{1,2},\ldots\ldots,y_{1,n}]$ and $y_2=[y_{2,1},y_{2,2},\ldots\ldots,y_{2,n}]$
-------------------------------------------------------------------------------------
Choose two crossover points a and b such that $1 \leq a \leq b \leq n$;
**Repeat**

      ***Permute*** $(x_a, x_b)$;
      a = a + 1;
      b = b − 1;
**until  a<b**

**Fig.8. Algorithm of RSM operator**

## 4.6  Insertion Method

We used the method of inserting elitism that consists in copy the best chromosome from the old to the new population. This is supplemented by the solutions resulting from operations of crossover and mutation, in ensuring that the population size remains fixed from one generation to another.

We would also like to note that the GAs without elitism can also be modeled as a Markov chain and Davis and Principe [38] proved their convergence to the limiting distributions under some conditions on the mutation probabilities [16]. However, it does not guarantee the convergence to the global optimum. With the introduction of elitism or by keeping the best string in the population allows us to show the convergence of the GA to the global optimal solution starting from any arbitrary initial population.

## 5.  NUMERICAL RESULTS AND DISCUSSION

Traveler Salesman Problem (TSP) is one the most famous problems in the field of operation research and optimization [1]. We use as a test of TSP problem the BERLIN52, witch has 52 locations in the city of Berlin (Fig. 9). The only optimization criterion is the distance to complete the journey. The optimal solution to this problem is known, it's 7542 m (Fig. 10).

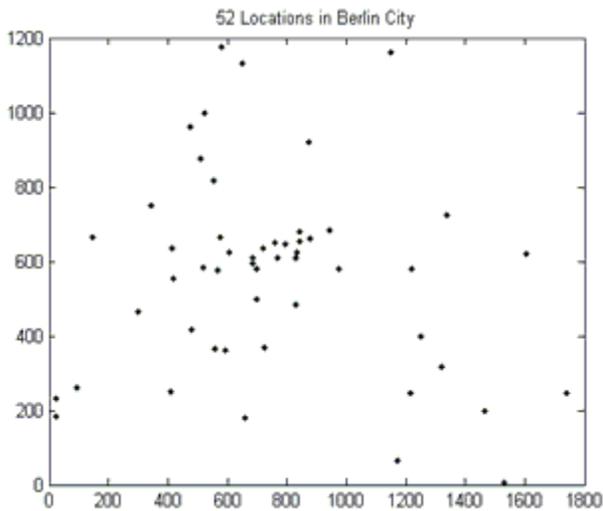

**Fig.9. The 52 locations in the Berlin city**

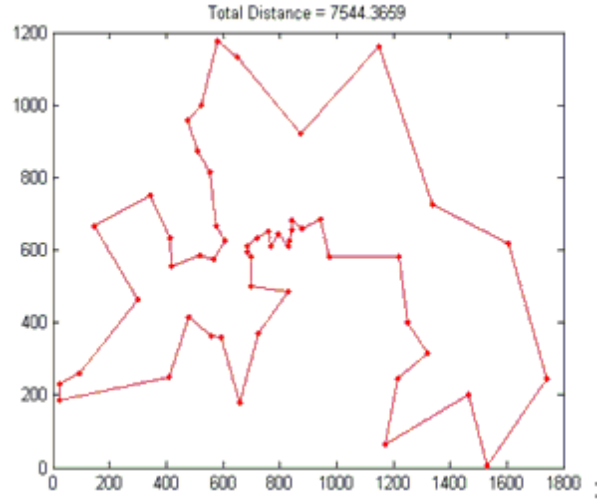

**Fig.10. The optimal solution of Berlin52**

## 5.1  Environment

The operators of the genetic algorithm and its different modalities, which will be used later, are grouped together in the next table (Table 10):

**Table 10. The operators used**

| Crossover operators | OX ; NWOX ; PMX ; UPMX ; CX |
|---|---|
| **Probability of crossover** | 1; 0.9 ; 0.8 ; 0.7 ; 0.6 ; 0.5 ; 0.4 ; 0.3 ; 0.2 ; 0.1 ; 0 |
| **Mutation operator** | PSM ; RSM |
| **Mutation probability** | 1; 0.9 ; 0.8 ; 0.7 ; 0.6 ; 0.5 ; 0.4 ; 0.3 ; 0.2 ; 0.1 ; 0 |

We change at a time one parameter and we set the others and we execute the genetic algorithm fifty times. The programming was done in C++ on a PC machine with Core2Quad 2.4GHz in CPU and 2GB in RAM with a CentOS 5.5 Linux as an operating system.

## 5.2  Results and Discussion

To compare statistically the operators, these are tested one by one on 50 different initial populations after that those populations are reused for each operator.

---

**Generate** the initial population $P_0$
i = 0
**Repeat**

      P'$_i$ = Variation ($P_i$);
      Evaluate (P'$_i$);
      P$_{i+1}$ = Selection ([P'$_i$, P$_i$]);
**Until** i<ltr

---

**Fig.11. Evolutionary algorithm**

To compare statistically the operators, these are tested one by one on 50 different initial populations after that those populations are reused for each operator. In the case of the comparison of crossover operators, the evolutionary algorithm is presented in Figure 11 which the operator of variation is given by the crossover algorithms and the selection is made by Roulette for choosing the shortest route.

Figure 12 shows the statistics of the experiments relating to the operators of crossover. It is interesting to note that the OX operator has not yet reached its shelf of evolution while the NWOX operator is on the quasi-shelf.

In addition, on average, NWOX does not always produce similar results, its standard deviation of the best of final individuals on 50 different initial populations is higher than all other operators, we can conclude that this operator is more much influenced by the initial population than its competitors.

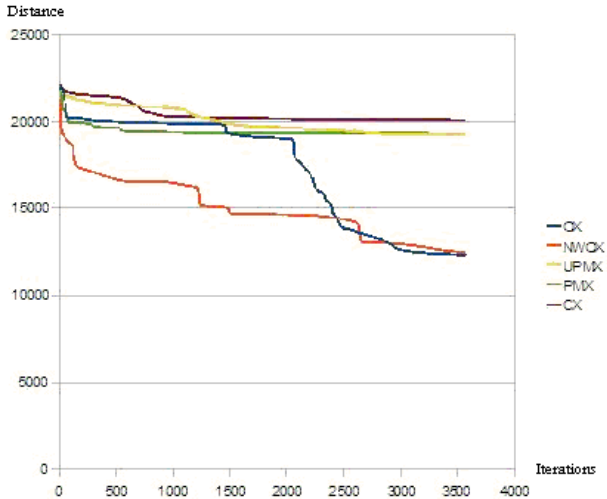

**Fig.12. Comparison of the crossover operators**

## 6. CONCLUSION

In this paper, the solution recombination, i.e. crossover operators in the context of the traveling salesman problem are discussed. These operators are known as playing an important role by developing robust genetic algorithms.

We implemented six different crossover procedures and their modifications in order to test the influence of the recombination operators to the genetic search process when applied to the traveling salesman problem. The following crossover operators have been used in the experimentation: the Uniform Crossover Operator (UXO), the Cycle Crossover (CX), the Partially-Mapped Crossover (PMX), the Uniform Partially-Mapped Crossover (UPMX), the Non-Wrapping Ordered Crossover (NWOX) and the Ordered Crossover (OX). The obtained results with *BERLIN52*, as a test instance of the TSP, show high performance of the crossover operators based on the creating and filling holes. The best known solution for the TSP instance BERLIN52 was obtained by using the OX operator.

According to the comparative study of the crossover operators mentioned, the development of innovative crossover operators for the traveling salesman problem may be the subject of the future research.